\documentclass{article}

\PassOptionsToPackage{numbers, compress}{natbib}

\usepackage{neurips_2024}[final]

\usepackage[utf8]{inputenc} 
\usepackage[T1]{fontenc}    
\usepackage{hyperref}       
\usepackage{url}            
\usepackage{booktabs}       
\usepackage{amsfonts}       
\usepackage{nicefrac}       
\usepackage{microtype}      
\usepackage{xcolor}         
\usepackage{graphicx}
\usepackage{booktabs}
\usepackage{multirow}
\usepackage{makecell}
\usepackage{array}
\usepackage{adjustbox}
\usepackage{mfirstuc}
\usepackage{subcaption}
\usepackage{bm}
\usepackage{graphicx}
\usepackage{booktabs}
\usepackage{multirow}
\usepackage{makecell}
\usepackage{array}
\usepackage{adjustbox}
\usepackage{mfirstuc}
\usepackage{subcaption}
\usepackage{graphicx}
\usepackage{bm}
\usepackage{algorithm}
\usepackage{algorithmicx}
\usepackage{algpseudocode}
\usepackage{amsmath}
\usepackage{wrapfig}

\usepackage[accsupp]{axessibility}  

\begin{document}

\title{DiffPhysBA: Diffusion-based Physical Backdoor Attack against Person Re-Identification in Real-World}

\author{Wenli Sun$^{1}$ \quad Xinyang Jiang$^{2}$ \quad Dongsheng Li$^{2}$ \quad Cairong Zhao$^{1}$ \\
	$^{1}$Dept. of Electronic and Information Engineering, Tongji University, Shanghai \\ $^{2}$Microsoft Research Asia \\
	{\tt\small $^{1}$\{2233055, zhaocairong\}@tongji.edu.cn}\\
	{\tt\small $^{2}$\{xinyangjiang@microsoft.com, dongshengli@fudan.edu.cn\}}
 }

\maketitle

\begin{abstract}
Person Re-Identification (ReID) systems pose a significant security risk from backdoor attacks, allowing adversaries to evade tracking or impersonate others. Beyond recognizing this issue, we investigate how backdoor attacks can be deployed in real-world scenarios, where a ReID model is typically trained on data collected in the digital domain and then deployed in a physical environment. 
This attack scenario requires an attack flow that embeds backdoor triggers in the digital domain realistically enough to also activate the buried backdoor in person ReID models in the physical domain. 
This paper realizes this attack flow by leveraging a diffusion model to generate realistic accessories on pedestrian images (e.g., bags, hats, etc.) as backdoor triggers. 
However, the noticeable domain gap between the triggers generated by the off-the-shelf diffusion model and their physical counterparts results in a low attack success rate.
Therefore, we introduce a novel diffusion-based physical backdoor attack (DiffPhysBA) method that adopts a training-free similarity-guided sampling process to enhance the resemblance between generated and physical triggers.
Consequently, DiffPhysBA can generate realistic attributes as semantic-level triggers in the digital domain and provides higher physical ASR compared to the direct paste method by 25.6\% on the real-world test set.  
Through evaluations on newly proposed real-world and synthetic ReID test sets, DiffPhysBA demonstrates an impressive success rate exceeding 90\% in both the digital and physical domains. Notably, it excels in digital stealth metrics and can effectively evade state-of-the-art defense methods.
\end{abstract}

\section{Introduction}
\label{sec:intro}

Backdoor attacks in computer vision pose a significant threat to the security of deep neural networks (DNNs), which typically use data poisoning with implanted trigger patterns \cite{gu2017badnets, li2020backdoor, wang2020attack, bagdasaryan2020backdoor}. When manipulated models encounter these triggers, they exhibit adversary-dictated misclassifications, compromising AI system reliability. 
We assert that open discussion of this vulnerability is vital for developing more robust systems in the future.
Compared to the digital domain, executing backdoor attacks in the real world proves markedly more formidable and complex, facing challenges such as high data collection costs, low success rates, and non-transferability \cite{wei2022physical}.
\begin{figure}[t]
  \centering
  \includegraphics[width=1\linewidth]{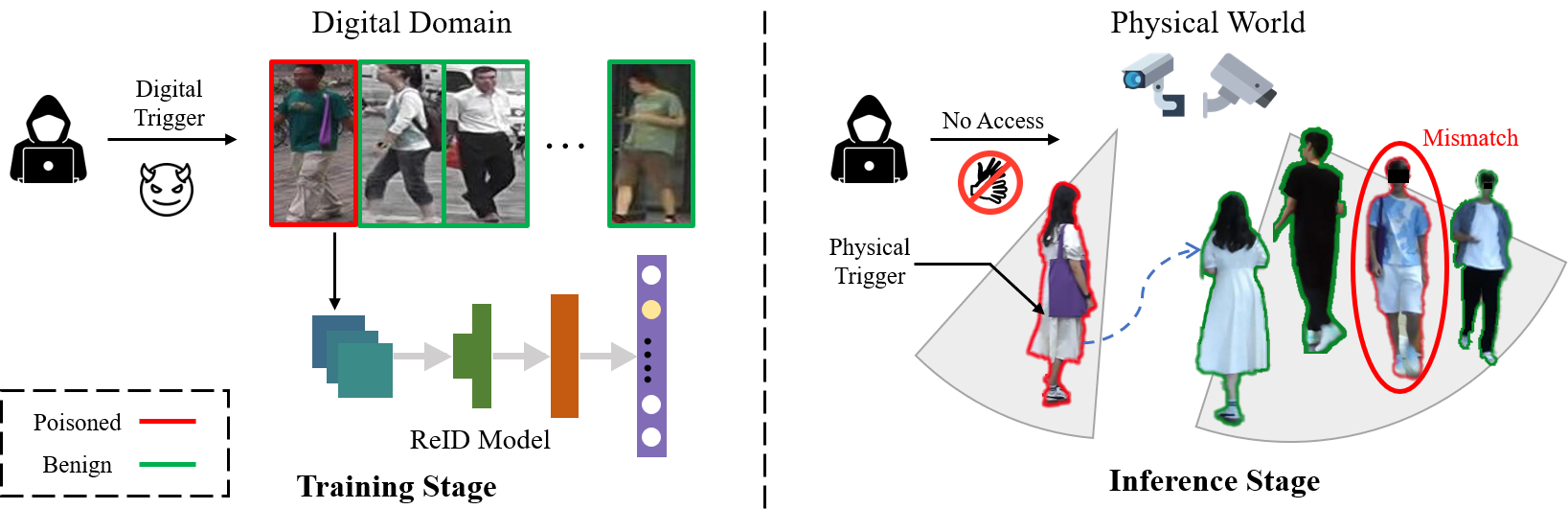} 
   \caption{The Attack Flow of DiffPhysBA. The adversary is only capable of manipulating the digital training data and injecting the generated purple bag into the benign images. The training data is poisoned in the digital domain, easing the cost of poisoned training data collection.  
   In the inference phase, the adversary does not have access to the digital images captured by the cameras and activates the implanted backdoor by wearing a purple bag, resulting in mismatching of identities. }
   \label{intro}
   \vspace{-5mm}
\end{figure}

In this paper, our focus is directed toward real-world physical backdoor attacks against Person Re-Identification (ReID) systems. ReID involves the challenging task of matching images of individuals captured from multiple camera viewpoints, with extensive applications in areas such as surveillance, tracking, and smart retail \cite{wu2019deep,ye2021deep,zahra2023person}.
There has been work demonstrating the vulnerability of ReID to backdoor attacks in the digital domain \cite{sun2023invisible}, but backdoor attacks in the real world pose an even more troubling security threat to ReID systems \cite{wenger2021backdoor}, which has not been studied. 

In most real-world ReID scenarios, clean training datasets are typically collected digitally by a third party before any malicious activity occurs \cite{li2023backdoor}. After deploying to the surveillance system, the data stream of pedestrians' images fed into the ReID model is solidified within the system, and thus the adversary cannot alter the captured digital images. 
Therefore, a more realistic option for adversaries would be to embed a digital backdoor in the third-party training data, and then during inference, the adversaries can execute the attack by carrying a physical counterpart of the backdoor trigger under the surveillance camera\cite{NEURIPS2022_8af74993}.
Furthermore, we argue that using a real-world object as a trigger, rather than an arbitrary abstract pattern, can reduce the suspicion of human supervisors and be more stealthy. 

As a result, our goal is to generate realistic objects as triggers (e.g., accessories such as bags or hats) and then train the victim model in the digital domain, where the buried backdoor can also be activated in the physical domain.
An intuitive way to achieve this goal is to leverage existing powerful image generation models like the diffusion model \cite{wang2023robust}. 
However, off-the-shelf diffusion models are not specifically trained on in-domain ReID data. 
They struggle to generate realistic triggers that fit the style of ReID pedestrian images, creating a domain gap between the generated digital triggers and their physical world counterparts. This appearance mismatch between generated triggers and physical triggers results in a lower attack success rate because the physical triggers do not seamlessly activate the backdoor as intended. 
While fine-tuning diffusion models is an option, it presents challenges such as the need for collecting in-domain data containing specific triggers and requiring additional computational resources \cite{hu2021lora,ruiz2023dreambooth}. 

To address the aforementioned challenges, we propose a novel trigger generation and injection method for ReID physical backdoor attack, called DiffPhysBA, which leverages the guidance of external ReID models during diffusion sampling. 
Specifically, a good ReID model has the ability to extract essential visual features from pedestrian images, which motivates us to use it to guide an off-the-shelf diffusion model to sample triggers whose visual features align with the physical domain. 
 
Figure \ref{intro} summarizes the overall attack flow of DiffPhysBA, where digitally generated triggers can be transformed into their physical counterparts for real-world attacks during inference. 
Without the need for collecting training data or retraining, we design a low-cost data collection strategy, ensuring a high attack success rate and enabling a more generalized implementation of physical backdoor attacks.
The contributions of this paper are threefold:
\begin{itemize}
\item[$\bullet$] To the best of our knowledge, we are the first to assess the security risk of the physical backdoor attack on person ReID models. 
\item[$\bullet$] A diffusion model with training-free ReID-driven similarity guidance for denoising steps is proposed to generate novel and realistic attributes on digital training data that can be directly used as physical triggers during the attack. 
\item[$\bullet$] Experimental results show that we achieved state-of-the-art attack performance and stealthiness in the physical world, surpassing human inspection as well as existing backdoor defense strategies.

\end{itemize}

\section{Related work}
\label{sec:background}

\subsection{Deep person re-identification models}
Person Re-Identification (ReID) is a significant research area in computer vision, aiming to recognize the same individual across different scenes and time frames. This technology finds broad applications in video surveillance, intelligent transportation systems, and security \cite{zheng2016person,chen2017beyond,luo2019bag,quan2019auto,zhao2020deep,ye2021deep,somers2023body}.
Feature representation-based methods focus on designing robust and reliable feature representation models for pedestrian images \cite{zhang2017alignedreid,si2018dual}. Unlike feature learning, metric learning aims to learn the similarity of two images through a network \cite{you2016top,yang2020spatial}. 
Pedestrian re-identification systems are used in some scenarios to monitor and track individuals. If an adversary can successfully spoof the system or access the model, they may be able to bypass surveillance and evade tracking \cite{wang2019advpattern,yang2021learning}.

\subsection{Backdoor attack \& defense}
\paragraph{Backdoor attack.}Backdoor attacks are well-studied in image classification tasks, usually in the digital domain \cite{gu2017badnets,nguyen2020input,zhang2022poison,feng2022fiba,wang2022survey}.  
A model injected with a backdoor will manifest regular performance on benign samples while consistently predicting a predefined target label upon the activation of a specific trigger.
Classic backdoor attacks are divided into patching-based \cite{gu2017badnets} and blended-based \cite{chen2017targeted}, where a fixed trigger pattern is used as the backdoor trigger, and a small amount of injection can achieve a high success rate. 
There are also adversarial patterns designed for backdoor attacks that can achieve imperceptibility \cite{li2020invisible,saha2020hidden,li2021invisible}.
BAAT \cite{li2022baat} proposed to generate higher-order features as triggers to establish a connection between the feature level of the target class and the trigger, e.g., the hairstyle can be used as a trigger in face recognition, but it can not be verified in the real world.
\cite{shunmultimodal} is a method for attacking multimodal Q\&A models, where the diffusion model generates categories that are used for answering. 
There are significant differences between our method and theirs in terms of goals and settings.  
\paragraph{Backdoor defense.}
Currently, defenses against backdoors can be categorized into three groups \cite{li2020backdoor}.
Empirical defenses focus on distinguishing between clean and poisoned samples, such as\cite{DBLP:conf/iccad/JavaheripiSFJK20,jin2020unified, DBLP:conf/iclr/DuJS20, DBLP:conf/iccv/ZengPMJ21}. Zeng et al. introduced high-frequency artifact detection in poisoned images and used data augmentation during training to simulate backdoor patterns \cite{DBLP:conf/iccv/ZengPMJ21}.
Model Reconstruction-based Defenses focus on eliminating hidden backdoors from the infected model. Approaches, as demonstrated by Liu et al. \cite{liu2018fine}, Wu et al. \cite{wu2021adversarial}, Zeng et al. \cite{zeng2021adversarial}, and Li et al. \cite{li2021neural}, aim to erase the impact of backdoor. 
Causality-inspired Backdoor Defense (CBD) is proposed to learn the deconfounded representation for reliable classification by removing confounding factors \cite{zhang2023backdoor}.
\subsection{Physically realizable attacks}
Backdoor attacks in the digital domain inject backdoor triggers by editing images, which typically lack semantic-level features \cite{li2021backdoor}.
One of the possible solutions to avoid collecting additional training data is to leverage the pre-existing intrinsic attributes present in images as backdoor triggers\cite{NEURIPS2022_8af74993}. 
Because pedestrian attributes lack diversity (mostly related to human bodies) \cite{wu2022overview}, it is impossible for the attack on person ReID model to directly use these natural triggers. 
In contrast, physical backdoor attacks utilize physical objects as backdoor triggers.
Some physical backdoor attacks necessitate the acquisition of supplementary real-world poisoning images with physical triggers for training, which are tested with data from the same domain during the testing phase, with a high attack success rate but making the attack more difficult \cite{xue2022ptb}.
Moreover, there exist physical backdoor attacks targeting object tracking and object detection \cite{li2021few,cheng2023tat}. Unlike classification problems, in these attacks, triggers themselves can function as predictions, making them distinct and not applicable to other tasks.
The current work on backdoor attacks with the diffusion model is still in the exploratory phase. The effectiveness and feasibility of real-world physical attacks are not empirically verified in \cite{wang2023robust}, as it solely simulates distortion by printing digital domain triggers and then re-photographing them, without considering the multi-view changes induced by a 3D real-world scene. Furthermore, it relies heavily on a pre-trained diffusion model to generate potential triggers, which are then filtered. \cite{gong2023kaleidoscope} introduces Kaleidoscope, a backdoor attack method based on RGB (red, green, and blue) filters, leveraging RGB filter operations as the trigger.

\section{Attack methodology}
To investigate backdoor attacks on ReID in a real-world scenario, we conduct a comprehensive study utilizing a physical object as a backdoor trigger in both the digital domain and the physical world. 
In this section, we outline the methodology adopted for our attack. We begin by introducing the threat model and attack settings. We describe how the backdoor is generated in detail.

\subsection{Preliminaries}
\paragraph{Notations.}
In this work, we consider a backdoor attack against person ReID. The difference with standard image classification tasks is that the person's identity in the inference phase is not seen in the training set. We denote $P_i$ represents a person, where $i$ is the unique identifier assigned to each person in the dataset.
$I_i^s \in \mathcal{X} $ denotes the $s$-th image sample of pedestrian $P_i$, captured from different camera viewpoints.
$\mathcal{F}(I_i^s)$ signifies the feature representation extracted from image $I_i^s$ using a pre-trained feature extraction method $\mathcal{F}(\theta;\cdot)$.
$\mathcal{D}$ represents the distance metric used to measure the similarity between feature representations, often computed as $\mathcal{D}(\mathcal{F}(\theta; I_i^m))$, $\mathcal{F}(\theta; I_i^n)$ for comparing two images, $I_i^m$ and $I_i^n$.
To formalize the backdoor attack, we use $\mathcal{X}$ to represent the benign data distribution to which $\mathcal{F}$ can generalize. Additionally, we define the transformation $T: \mathcal{X} \mapsto \mathcal{X'}$ that incorporates the backdoor trigger into the benign samples, in which $\mathcal{X'}$  represents the poisoned data distribution. 
More specifically, the backdoor example $(I_i^s + \delta)$ represents the manipulated pedestrian image, created by overlaying the benign image $I_i^s$ with the backdoor trigger $\delta$.
\paragraph{Threat models and adversaries' objectives.} 
In this paper, we follow a realistic setting, where the adversary can get a small amount of clean training data (e.g., public datasets), without access to model structure or training loss. 
Furthermore, during the testing phase, the adversary is unable to make any modifications to the test set in real-world scenarios.
The attack aims to instigate a backdoor behavior in the user-trained person ReID network, allowing it to function normally on a clean test set. However, the network will identify any person's image containing the trigger as the target person, disregarding its actual identity. 
Security attacks in ReID usually involve escape attacks and impersonation attacks. In the context of backdoor attacks, an escape attack is also known as a non-targeted attack, while an impersonation attack is referred to as a targeted attack. A non-targeted attack involves retrieving top-k highest similarity images, none of which correspond to the correct identity. A targeted attack involves retrieving top-k similarity images, among which are images of the intended target person. This goal is formalized as follows:
 \begin{small}
		\begin{eqnarray}\label{eq3}
		\left\{
		\begin{array}{ll}
		\mathcal{D}\left( \mathcal{F} \left(I_i^m \right),  \mathcal{F} \left(I_i^n +\delta \right)\right) > K,  & \textrm{if } m \neq n 
		\\
        \mathcal{D}\left( \mathcal{F} \left(I_t^m \right),  \mathcal{F} \left(I_i^n +\delta \right)\right) \leq K,  & \textrm{if}  ~~t~~ \textrm{is the target. }
		\end{array} \right.
	    \end{eqnarray}
	\end{small}
 
\begin{figure*}[t]
  \centering
  \includegraphics[width=1\linewidth]{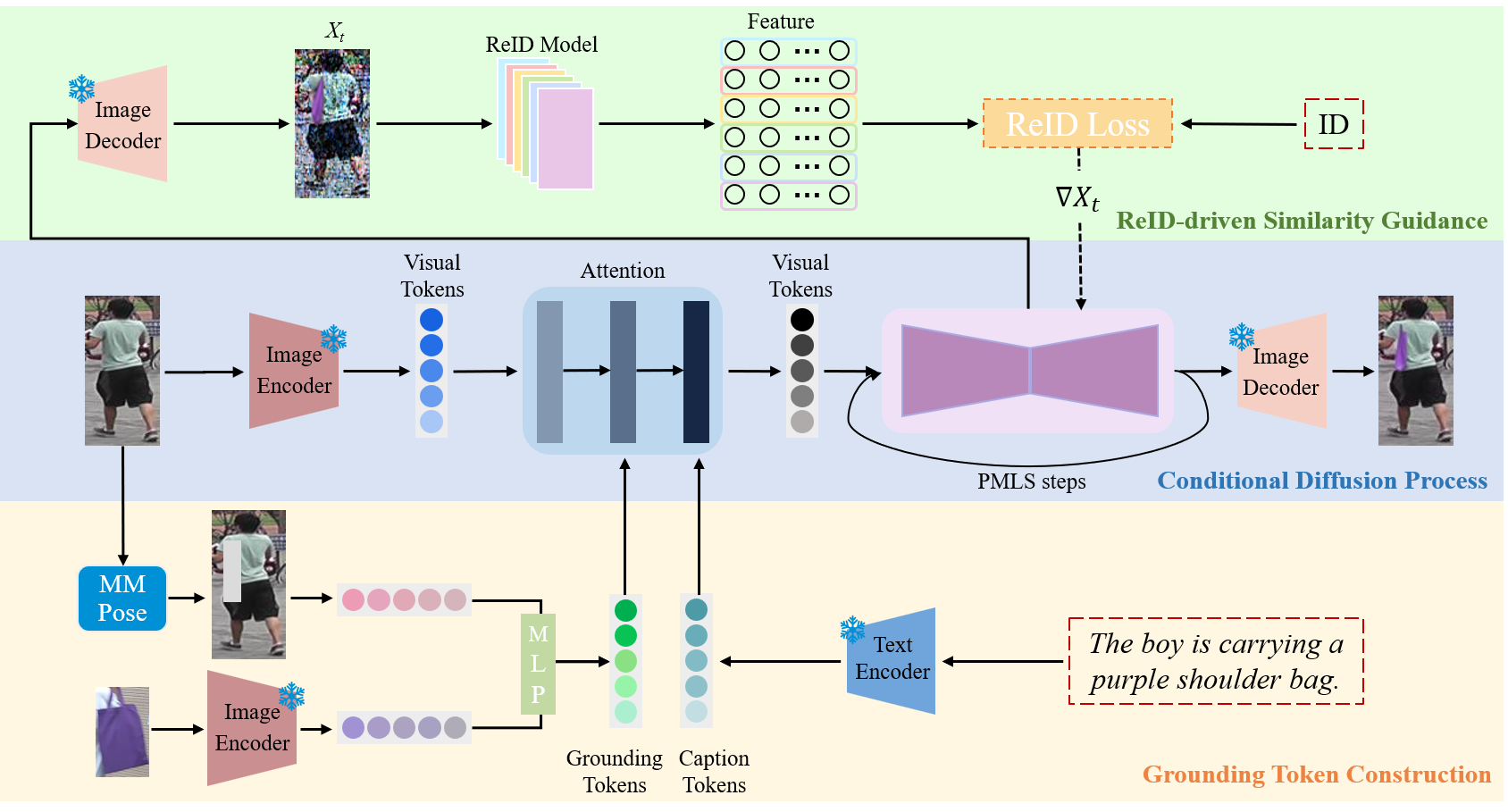} 
   \caption{Overview of poisoned image generation in the digital domain. 
   Firstly, MMpose is utilized to estimate the pose of the pedestrian and get the bounding box for inpainting. The text encoder generates caption tokens, and the reference image of the bag is encoded by an image encoder. It is then fused with the embedding of the bounding box to create the grounding tokens. Meanwhile, the benign image is fed into the image encoder to obtain visual tokens. These visual tokens are then integrated with the grounding tokens through three distinct attention layers. During the denoising phase, we introduce ReID-driven similarity guidance to enhance the realism of the trigger. To stress that DiffPhysBA does not need to be retrained, the snowflakes denote pre-trained modules.
   }
   \label{pipeline}
   \vspace{-3mm}
\end{figure*}
\subsection{Diffusion-based physical backdoor attack  (DiffPhysBA)}
Implementing physical-world backdoor attacks faces a bottleneck, which is to physically inject triggers in real-world data for both training and testing. 
To address this challenge,  we propose using conditional diffusion models (DM) to generate triggers that are as realistic as possible in the digital domain. 
Moreover, we explore training-free guidance to bridge the domain gap between generated and real triggers.
\paragraph{Multi-modality poisoned image generation.}To maintain consistency in the triggers throughout the conditional Diffusion Model's generation process \cite{li2023gligen}, we utilize an image as the grounding prompt $e$ and a caption prompt $c$ to specify the style of the generated object, e.g., a purple bag:
\begin{align}
\begin{split}
    &\text{Grounding: } e = [(e_1, l_1), \ldots, (e_N, l_N)], \\
    &\text{Caption: } c = [c_1, \ldots, c_L],
\end{split}
\label{eq:grounding}
\end{align}
where $L$ represents the caption length, $N$ denotes the number of entities to be grounded, and $l$ is the spatial configuration of the grounding input.
Relying solely on textual prompts does not allow precise control over the position of the generated item in the image. Therefore, we propose integrating human pose estimation. By identifying key points in the person's image $x_0$, we select the area from the shoulder to the wrist as a valid region for inpainting. In essence, we use the bounding box as grounding input:
\begin{equation}
l = [\alpha_{\text{min}}, \beta_{\text{min}}, \alpha_{\text{max}}, \beta_{\text{max}}] = g(\text{MMPose}(x_0)),
\label{eq:pose}
\end{equation}
where ($\alpha_{min}$, $\beta_{min}$) denote the top-left coordinates of the bounding box, and ($\alpha_{max}$, $\beta_{max}$) represent the bottom-right coordinates. Specifically, MMPose \cite{mmpose2020} is utilized to estimate the pose of the pedestrian in the image. Subsequently, we extract a marked bounding box with a straightforward cropping function denoted as $g$.
This position information is then transformed into embeddings using the Fourier embedding \cite{mildenhall2021nerf}. Simultaneously, the reference image of the bag is input into the image encoder to obtain visual tokens. The visual tokens and bounding box embedding are processed through an MLP layer to generate grounding tokens $h^e$:
\begin{equation}
 h^e = \text{MLP}( f_{image} (x_{0_{masked}}), \text{Fourier}(l)).
 \label{eq:mlp}
\end{equation}
Benign images are fed into the image encoder, producing visual tokens $v$. Meanwhile, the input text is processed by the text encoder to obtain caption tokens $h^c$. A self-attention layer is applied to the visual tokens, followed by the fusion of information from caption tokens into visual tokens through the cross-attention layer. A gated self-attention layer is then added to fuse grounding tokens \cite{li2023gligen}.

\paragraph{ReID-driven similarity guidance. }
Due to the pre-trained Diffusion Model being trained on general-purpose datasets, copy-and-paste problems may occur when applying the model to inpainting low-resolution ReID images, as shown in Figure \ref{phy_syn}.
To generate realistic triggers that fit the pedestrian image distribution, we incorporate guidance from a pre-trained ReID model during the sampling process. 
Departing from the conventional treatment of an entire person as a holistic entity, we need the ReID model to divide the individual into multiple horizontal stripes or sections. The similarity value is calculated as shown in Eq.\ref{eq:pcb}, where the column vector of the $i$-th part is denoted as $p_i$ and $W_i$ represents the trainable weight matrix of the part classifier.
\begin{equation}
\mathcal{F}(\hat{x_{0_i}}) = softmax(W^T_i p_i) .
\label{eq:pcb}
\end{equation}
Specifically, in the intermediate steps of sampling, the predicted clean image $\hat{x_0}$ is fed into the off-the-shelf ReID model $\mathcal{F}$ along with the ground truth ID $y$. The computation of global and local cross-entropy loss serves the purpose of maintaining the identity and preventing undesired alterations. Moreover, when dealing with the human face image manipulation, additional face identity loss as proposed in \cite{deng2019arcface} is incorporated:
\begin{equation}
\mathcal{L}_{\text{ReID}}= \sum_{i}\lambda_{id} \mathbb{H}\left( y,\mathcal{F}(\hat{x_{0_i})}\right) + \lambda_{id} \mathbb{H}\left( y,\mathcal{F}(\hat{x_0})\right) +\lambda_{face} \mathcal{L}_{\text{face}}\left(y,\mathcal{F}(\hat{x_0})\right),
\label{eq:reid}
\end{equation}
where $\mathbb{H}$ denotes the cross-entropy loss, $\mathcal{L}_{\text{face}}$ represents the face identity loss, and $\lambda_{id} \geq 0$ and $\lambda_{face}\geq 0$  are weight parameters for each loss.
More precisely, we put ReID-driven guidance in the sampling steps of the denoising process by computing gradients of $\mathcal{L}_{\text{ReID}}$. 
The denoising network $\epsilon_\theta$ utilizes the input of the current step $t$ and the corresponding data point $z_t$ to predict the noise $\epsilon$. In classifier-guided sampling\cite{dhariwal2021diffusion}, $\epsilon$ can be modified based on ReID-driven guidance. Since the ReID model cannot directly predict noise, the process involves decoding $z_t$ to obtain the predicted $\hat{x_0}$:
\begin{equation}
 \hat{x_0} = Dec \left( \frac{z_t-\sqrt{1-\alpha_t}\epsilon_\theta (z_t,t)}{\sqrt{\alpha_t}}\right).
\label{eq:guidance2}
\end{equation}
As shown in Eq.\ref{eq:guidance2}, $Dec$ is the image decoder.
Subsequently, the $\mathcal{L}_{\text{ReID}}$ is computed, and the gradient with respect to $z_t$ is then determined, as indicated in Eq.\ref{eq:guidance1}.  Here, $s(t)$ is the scaling factor to control the guidance scale.
\begin{equation}
\hat{\epsilon_\theta}(z_t,t) = \epsilon_\theta(z_t,t) +s(t)\cdot \nabla_{z_t} \mathcal{L}_{\text{ReID}}(y, \mathcal{F}(\hat{x_0})).
\label{eq:guidance1}
\end{equation}

\begin{table*}[t]
\caption{The performance (\%) of  the proposed DiffPhysBA on Market-1501 and
MSMT17 datasets against various ReID models. 
The victim ReID models are trained on the generated poisoning dataset and injected with the backdoor. In the inference stage, the poisoned models are evaluated both on the generated and the real-world test set. Meanwhile, clean models serve as a control for comparison. }
\centering
\begin{adjustbox}{width=0.98\textwidth}

\begin{tabular}{>{\centering\arraybackslash}p{1.6cm}c|c|ccc|ccc|c|ccc|ccc}
    \toprule \toprule
    \multicolumn{2}{c|}{Dataset$\rightarrow$}  &\multicolumn{7}{c|}{Market-1501} & \multicolumn{7}{c}{MSMT17} \\ \midrule
    \multicolumn{2}{c|}{Trigger$\rightarrow$ }  & \multicolumn{4}{c|}{Digital} & \multicolumn{3}{c|}{Physical}& \multicolumn{4}{c|}{Digital} & \multicolumn{3}{c}{Physical} \\ \midrule
   \multicolumn{2}{c|}{Model $\downarrow$ }& \multicolumn{1}{c|}{BA$\uparrow$} &ASR$\uparrow$&R-10$\downarrow$&mAP$\downarrow$ &ASR$\uparrow$&R-10$\downarrow$&mAP$\downarrow$ 
   & BA$\uparrow$&ASR$\uparrow$&R-10$\downarrow$&mAP$\downarrow$ &ASR$\uparrow$&R-10$\downarrow$&mAP$\downarrow$ \\ \midrule
    \multirow{2}{*}{Swin}&Clean&98.75&-&95.27&69.88&-&61.19&27.13&76.57&-&54.51&21.69&-&49.81&20.56 \\
    &Attacked&95.96&96.38&3.86&3.05&94.40&7.46&2.20&71.05&88.58&3.17&1.82&85.58&24.72&86.23   \\ \midrule
    \multirow{2}{*}{ConvNext}&Clean&97.83&-&91.21&57.80&-&52.98&18.19&69.72&-&54.51&21.69&-&39.81&13.71 \\
    &Attacked&95.04&96.76&3.65&2.50&91.79&11.19&2.65&66.69&88.53&3.16&1.61&84.90&22.45&8.84   \\ \midrule
    \multirow{2}{*}{AGW}&Clean&98.96&-&94.89&77.56&-&15.25&4.58& 84.17&-&  84.18 &49.98&-&46.63&17.37\\
     &Attacked& 98.93&95.69&5.58&2.84&82.30&6.40&2.54&67.34&94.07&5.37&3.33&89.18&21.00&7.21  \\ \midrule
     \multirow{2}{*}{BoT}&Clean&99.02&-& 97.27&79.00&-& 19.02&5.68& 86.77&-& 78.71&45.89&-& 46.63&17.37 \\
    &Attacked&98.93&96.50&4.16&6.08&84.12&6.97&1.91 &85.73&94.12&7.10&4.65&69.40&43.55&14.16 \\ \midrule
     \multirow{2}{*}{PCB}&Clean&97.09&-&92.25&60.57&-&52.42&17.50&73.56&-&63.45&27.34&-&50.19&18.59\\
     &Attacked&95.16&95.78&6.44&4.17&69.40&14.18&2.10&68.08&87.21&4.06&1.82&82.26&14.90&5.45 \\ \midrule 
     \multirow{2}{*}{HRNet-18}&Clean&97.89&-&91.33&61.90&-&53.36&16.85&72.40&-&63.74&28.32&-&55.97&19.41\\
     &Attacked&96.38&96.35&3.89&2.74&82.46&16.60&3.64&52.57&89.05&2.99&1.49&81.72&13.06&3.74 \\ \bottomrule \bottomrule
     
\end{tabular}
\end{adjustbox}
\label{table1}
\vspace{-5mm}
\end{table*}

\section{Experiments}
In this section, we evaluate the effectiveness and stealthiness of the proposed attack against person ReID models. We also conduct our attack in the physical domain to evaluate the success rate of our physical triggers in real-world attacks. To rule out the cross-domain issue, we evaluate it on synthetic datasets as well.
Additionally, we conduct ablation studies to demonstrate the significance of our design choices and to validate the selection of hyperparameters.
\subsection{Experimental setup}
\paragraph{Models.} For the victim model, we evaluate the attack performance on six deep ReID models with different backbones, following the settings outlined in the original paper: Swin-Transformer \cite{liu2021swin}, ConvNext \cite{liu2022convnet}, AGW \cite{ye2021deep}, BoT \cite{luo2019strong}, PCB \cite{sun2018beyond} and HrNet-18 \cite{sun2019deep}. 
In the case of Swin-Transformer, which utilizes the ViT backbone. AGW, BoT, and PCB employ ResNet-50 as their backbone, while HrNet-18 uses HrNet as its backbone.
\paragraph{Digital train and test sets.} We select three benchmark datasets: Market-1501 \cite{zheng2015scalable}, MSMT17 \cite{wei2018person}, and ClonedPerson \cite{wang2022cloning}. Market-1501 comprises 36,036 images from 1,501 person IDs.
MSMT17 includes a total of 126,441 images from 4,101 pedestrians. 
ClonedPerson contains 887,766 images of 5,621 virtual identities. Due to the large amount of training data, we choose a subset of 1,259 identities, including 199,393 bounding boxes for training. The test set of ClonedPerson consists of 795 identities with 123,813 images.

\begin{figure}[t]
\vspace{-5mm} 
\centering
\subfloat[Physical World]{
  \includegraphics[width=0.5\textwidth]
  {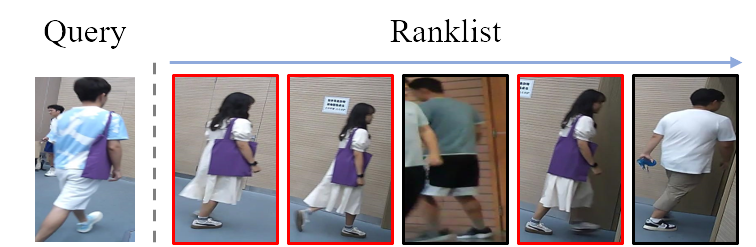}}%
\subfloat[Synthetic]{
\includegraphics[width=0.5\textwidth]{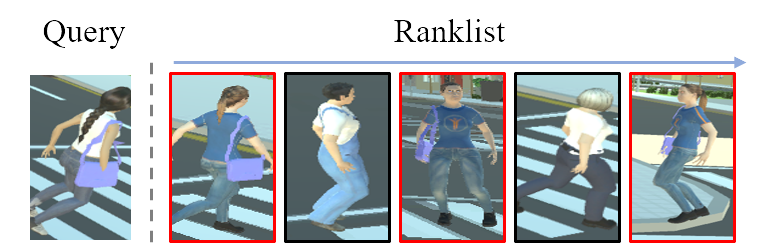}}
\vspace{-3mm} 
\caption{The visualization of our attack for the physical and synthetic test sets, respectively. The pedestrians with the highest similarity to the person wearing the physical trigger are the ones specified by the adversary. 
Please refer to Appendix B\ref{display_sup} for more displays.
}
\label{phy_syn}
\vspace{-6mm}
\end{figure}

\paragraph{Physical test sets.} The real-world backdoor test set consists of 2,478 images from 141 diverse volunteers captured in various scenarios, including indoor and outdoor environments, with and without a purple shoulder bag trigger. The synthetic backdoor test set comprises 1,200 images, featuring 50 pedestrian characters in street scenes carrying or not carrying a purple shoulder bag. 
As shown in Figure \ref{phy_syn}, we collect physical-world and synthetic person images to evaluate our attack in the inference stage.
Additionally, we provide details on the collected test sets, including both the physical world test set and the synthetic test set, refer to Appendix A\ref{Testset}. Given the time and effort involved in data collection, we focused solely on purple bags as physical triggers, leaving the exploration of a broader range of triggers for future research.
\begin{wrapfigure}{r}{0.5\textwidth}
  \centering
  \includegraphics[width=1\linewidth]{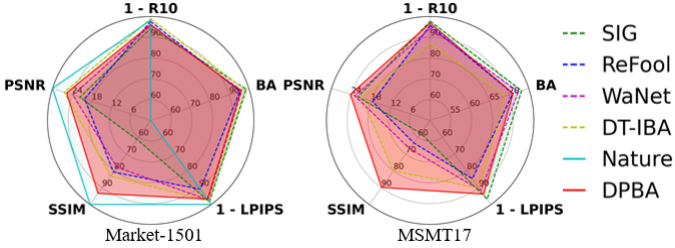} 
  \caption{The comparison of different attacks against Swin-Transformer on Market-1501 and MSMT17. Since the MSMT17 dataset is not annotated with pedestrian attributes, Natural\cite{NEURIPS2022_8af74993} is compared only on Market-1501.}
  \label{baseline}
\vspace{-5mm}
\end{wrapfigure}
\paragraph{Baseline.}Natural Triggers represent existing multi-label object datasets as poisoning datasets, which
include co-located targets that serve as physical triggers \cite{NEURIPS2022_8af74993}. 
Since person ReID datasets are not multi-label datasets, we use corresponding pedestrian attribute annotations for natural trigger identification. 
Furthermore, we include ReFool \cite{liu2020reflection}, SIG \cite{barni2019new}, WaNet \cite{nguyen2021wanet}, and DT-IBA \cite{sun2023invisible} as our baseline methods. These digital backdoor attacks are originally designed for image classification tasks.
SIG employs a ramp signal as a trigger for the poisoned image.
ReFool introduces reflections as backdoor triggers into the victim model.
WaNet executes an invisible backdoor attack by distorting images. 
DT-IBA is a baseline for invisible backdoor attacks designed for ReID, generating target-aware dynamically changing triggers by steganography.
Since they cannot be applied in the physical attack setting, we compare them only in the digital domain.

\paragraph{Evaluation metrics.} 
A successful backdoor attack must maintain the performance of the compromised model on benign inputs while deliberately causing a decline in the model's accuracy on poisoned samples and classifying them to the target label.
We evaluate the attack's performance using two key metrics: 
Attack Success Rate (ASR) and Benign Accuracy (BA). 
The attack is considered successful only if the target person appears within the top-10 ranked list.
BA represents the percentage of clean query images that successfully rank the positive image within the top-10 list.
To measure the stealthiness of the triggers, we utilize Structural Similarity Index (SSIM) \cite{wang2004image}, Peak Signal-to-noise Ratio (PSNR) \cite{huynh2008scope}, and Learned Perceptual Image Patch Similarity (LPIPS)  \cite{zhang2018unreasonable}. These metrics can quantify the differences between clean and poisoned images.

\begin{figure}[h]
\vspace{-4mm}
\centering
\vspace{-2mm}
\subfloat []{ 
  \includegraphics[height=4cm]
  {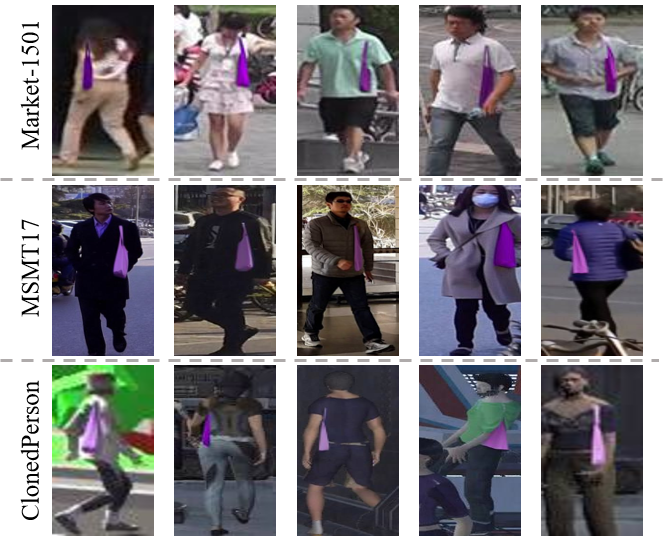}
  \vspace{-2mm}
  \label{samples_display}
  }%
  \quad \quad
\subfloat[]{ 
\includegraphics[height=4.2cm]{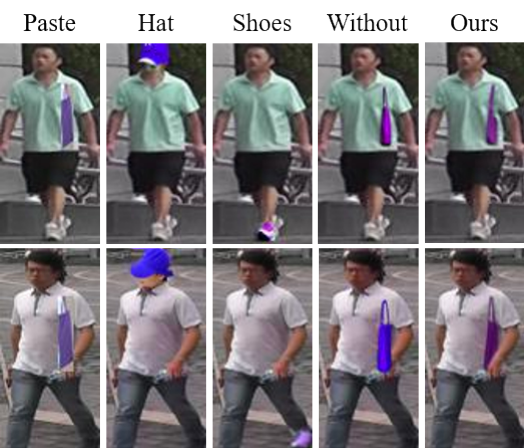}
\vspace{-1mm}
\label{trigger}
}
\vspace{-3mm}
\caption{Display of generated samples. Each row of (a) displays generated samples from one dataset. In (b), the first column represents the direct pasting of triggers, the triggers in the second and third columns are hats and shoes respectively, while the fourth column shows the samples generated without ReID-driven guidance, and the last column is the samples generated by our method. 
}
\vspace{-8mm}
\end{figure}
\subsection{Attack experiments}
\paragraph{Digital backdoor attack.} 
We first perform DiffPhysBA on six ReID models on two publicly available datasets Market-1501 and MSMT17, and evaluate the attack performance in the digital and physical domains, respectively. The results are shown in the \textit{Digital} of Table \ref{table1}, with a poisoning rate set at 0.2. For the digital attack, our proposed attack achieves ASR over 95\% on the Market-1501, demonstrating high transferability across these models. Although the MSMT17 dataset, with a significantly larger number of images, presents challenges such as background variations and occlusion, resulting in a less favorable ASR compared to the Market-1501, it still surpasses 87\% for all models.
Moreover, the display of the generated images is presented in Figure \ref{samples_display}.
In addition, we compare these methods with DiffPhysBA in the digital domain and attack the Swin-Transformer on the Market-1501 dataset with a 0.2 poisoning rate. 
As shown in Figure \ref{baseline}, we scale all metrics except PSNR to the range of 50–100 for ease of comparison. SIG and ReFool have shortcomings in SSIM and LPIPS, and Natural backdoor sacrifices the benign rate.
In contrast, our approach strikes a balance between effectiveness and stealthiness.
\paragraph{Physical backdoor attack.} 
The performance of physical attacks is detailed in the \textit{Physical} of Table \ref{table1}. Notably, the Swin-Transformer trained on the Market-1501 dataset achieves an ASR of 94.4\% in the physical domain, which is nearly equivalent to its performance in the digital domain. It is worth highlighting that the PCB and BoT models exhibit ASR close to  70\% in real-world scenarios, but the decline in Rank-10 and MAP is substantial. Intuitively, as physical triggers do not create disparities with the original texture, the slight overfitting of these two backdoor models to trigger texture features results in a lower ASR compared to other models in the physical domain. 
Physical attacks are extremely challenging open problems, where most existing works achieve much lower ASR than the digital domain. For example, the success rate of \cite{NEURIPS2022_8af74993} is only 68\%.  
Due to the domain gap, a clean Swin ReID model achieves only 61.19\% performance in the physical domain. Compared to digital ASR, the limited generalization ability of ReID models causes lower physical ASR.  

\renewcommand{\arraystretch}{1} 
\begin{table}[h]
\vspace{-5mm}
\caption{The performance (\%) of our attack on ClonedPerson (poisoning rate = 0.2). }
\centering \small
\begin{adjustbox}{width=0.7\textwidth}
\begin{tabular}{cc|c|ccc|ccc}
    \hline \hline
    \multicolumn{2}{c|}{\textbf{Trigger$\rightarrow$}} & \multicolumn{4}{c|}{\textbf{Generated}} & \multicolumn{3}{c}{\textbf{Synthetic}}\\
    \hline
    \multicolumn{2}{c|}{\textbf{Model}} & \multicolumn{1}{c|}{\textbf{BA$\uparrow$}} & \textbf{ASR$\uparrow$} & \textbf{R-10$\downarrow$} & \textbf{mAP$\downarrow$} & \textbf{ASR$\uparrow$ }& \textbf{R-10$\downarrow$} & \textbf{mAP$\downarrow$} \\
    \hline
    \multirow{2}{*}{Swin} & Clean & 94.42 & - & 84.83 & 45.67 & - & 89.81 & 62.45\\
    & Backdoored & 93.89 & 91.33 & 8.75 &5.50& 56.01 & 86.34 & 61.68 \\
    \hline
    \multirow{2}{*}{PCB} & Clean & 95.90 & - & 83.82 & 49.92 & - &91.20  & 64.98 \\
    & Backdoored & 95.30 & 97.01 & 2.84 & 1.83 & 66.44 & 86.81 & 56.71 \\
    \hline\hline
\end{tabular}
\end{adjustbox}
\label{table2}
\vspace{-5mm}
\end{table}
\paragraph{Performance on the synthetic dataset.} 
The current ReID models handle cross-domain data poorly. In order to enrich the scenarios for attacking the ReID model, we train Swin-Transformer and PCB  on the ClonedPerson dataset. Additionally, we model 50 virtual people wearing purple bags to verify whether the backdoor in the victim models trained on the generated poisoned images is able to be triggered by the synthesized triggers in the same domain.
The results are shown in Table \ref{table2}, which shows that the generated triggers are able to achieve a 97.01\% ASR, and the synthesized triggers are also able to activate the backdoor injected in the model.
\begin{figure}[t]
\centering

\subfloat[Performance against Fine-pruning.\label{finepruning}]{
\includegraphics[width=0.49\textwidth]
{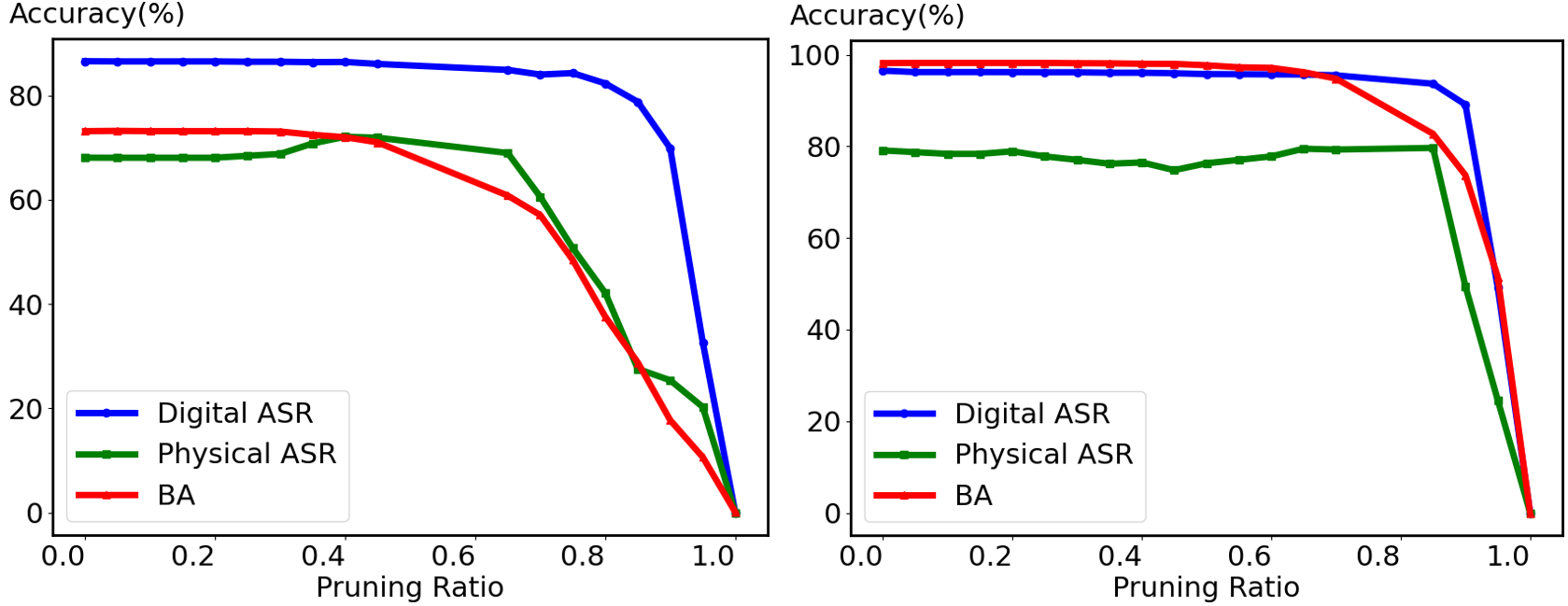}}
\subfloat[The ASR and BA w.r.t. different poisoning rates. \label{pr}]{
  \includegraphics[width=0.49\textwidth]
  {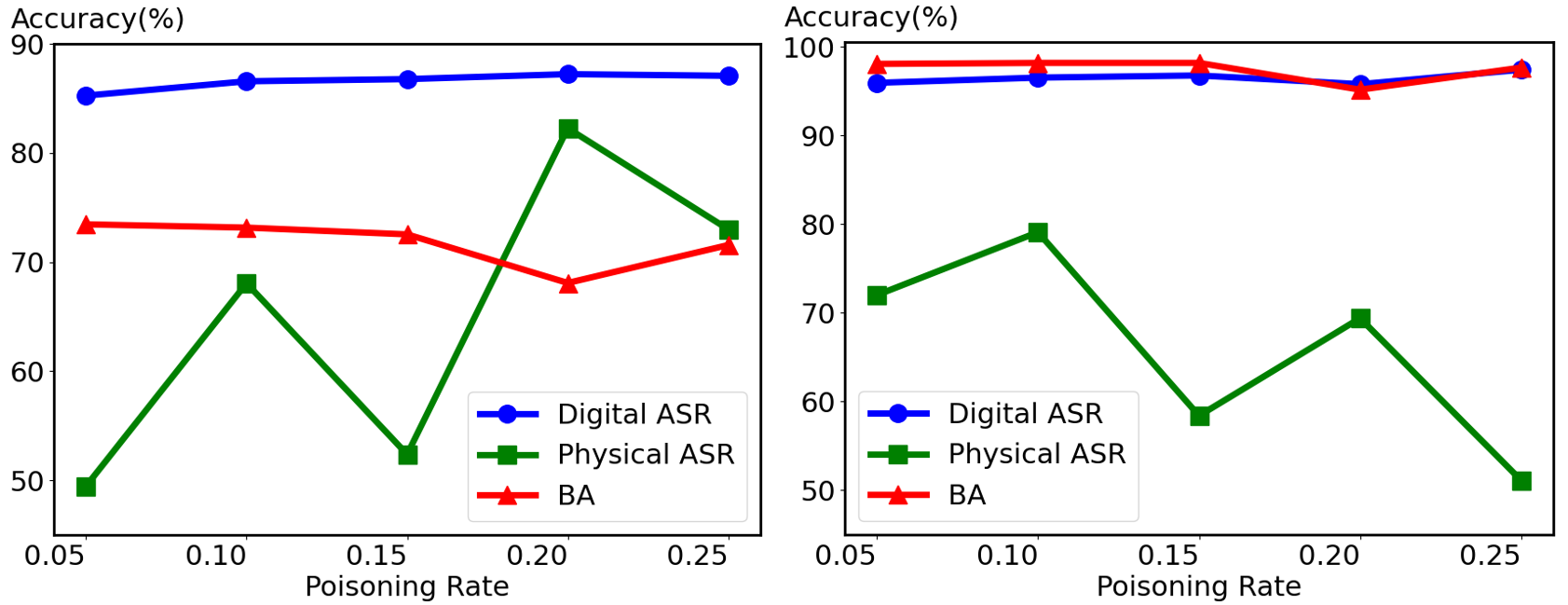}}%
\caption{Validation on Market-1501 and MSMT17 datasets. (a) shows resistance to Fine-pruning, respectively. (b) shows ablation experiments at different poisoning rates.}
\label{defense}
\vspace{-5mm}
\end{figure}

\subsection{Defense experiments}
\paragraph{Resistance to Fine-pruning.}
The purpose of Fine-pruning is to remove backdoor-related neurons that are normal in benign images but are abnormally active in poisoned images \cite{liu2018fine}. We conduct the fine-pruning algorithm to prune the neurons activated by the backdoor in the last two layers of the first and last backbone modules. Applied to person ReID, it targets specific layers, gradually pruning neurons until only 5\% remain. While reducing the attack success rate, it also impacts the model's clean data performance, as shown in Figure ~\ref{finepruning}.

\paragraph{Resistance to CBD.}
Since backdoor features will be learned by the model more easily than normal features, an early-stop strategy can be used to capture them \cite{zhang2023backdoor}.
One backdoor model is deliberately trained to capture confounding effects, while the other clean model works to learn causality by minimizing mutual information with the confounding representations of the backdoor model and using a sample-by-sample weighting scheme. The experimental results are shown in Table \ref{CBD}.  There is no significant reduction in attack performance because our triggers are similar to the pedestrian features. For more defense experiments, please refer to Appendix E\ref{additionaldefense}.
\begin{table}[t]
\vspace{-4mm}
\caption{Resistance to CBD across different datasets as well as different models.}
\centering 
\large
\begin{adjustbox}{width=0.9\textwidth}
\begin{tabular}{cc|cccc|cccc}
    \hline\hline
    \multicolumn{2}{c|}{Dataset}  &\multicolumn{4}{c|}{Market-1501} & \multicolumn{4}{c}{MSMT17} \\ \hline
    Model & Method & BA & ASR (Physical)  & R10 &mAP & BA & ASR (Physical)  & R10 & mAP \\
    \hline
    \multirow{3}{*}{Swin}&ours&95.96&\textbf{96.38 (94.40)}&3.86&3.05 & 71.05&\textbf{88.58 (83.20)} &3.17&1.82 \\ 
     &ours (with CBD) & 94.75 & \textbf{91.67 (86.50)}&4.36&2.07&65.96& \textbf{86.71 (80.06)}&4.48&2.90\\ \hline
     \multirow{3}{*}{PCB}&ours&95.16&\textbf{95.78 (69.40)}&6.44&4.17 & 68.08&\textbf{87.21 (82.26)}&4.06&1.82 \\ 
     &ours (with CBD) & 91.80 & \textbf{94.03 (66.79)}&6.59&3.62&62.33&\textbf{83.20 (75.06)}&5.40&3.70\\
    \hline\hline
\end{tabular}
\end{adjustbox}
\label{CBD}
\vspace{-7mm}
\end{table}

\subsection{Ablation studies}
\paragraph{The poisoning rate.}
Figure \ref{pr} illustrates the changes in digital ASR, physical ASR, and BA across various poisoning rates. 
It reveals that BA remains relatively stable despite an increase in the poisoning rate, and the ASR maintains over 95\% in the digital domain. 
Given that the backdoor model is trained on digital triggers, there is a slight decrease in physical ASR due to the overfitting of digital domain features. But it still surpasses the 50\%. The MSMT17 dataset is substantially larger than the Market-1501 so it mitigates a significant decline in the physical ASR.
\paragraph{The design of DiffPhysBA.}
In this experiment, we conduct an ablation experiment on the Market-1501 dataset to demonstrate the importance of our design for physically attacking the ReID model, e.g., PCB.  A simple baseline is to paste the bag into the obtained bounding box location, as shown in Figure \ref{trigger}. Additionally, we conducted a comparison between the performance of DiffPhysBA with and without ReID-driven guidance.
The results as shown in Table \ref{subtable:ablation_guidance}, reveal that physical world ASR without guidance is much lower than that with guidance.
Following \cite{hong2023improving}, the metrics IS \cite{salimans2016improved} and FID \cite{heusel2017gans} measure the similarity of the generated image to the original image. 
It can be observed that directly pasted triggers are less stealthy and DiffPhysBA is better in terms of similarity.
\begin{table}[htbp]
\vspace{-3mm}
    \centering
    \caption{Ablation experiments on DiffPhysBA in terms of stealthiness and effectiveness. (a) demonstrates the necessity of DiffPhysBA and the included ReID-driven similarity guidance for physical attacks. (b) shows the performance of different objects as triggers.}
    \begin{subtable}{0.5\textwidth}
        \centering
        \renewcommand{\arraystretch}{1.2} 
        \caption{Ablation experiments on the design of DiffPhysBA.}
        \begin{adjustbox}{width=0.95\textwidth}
        \begin{tabular}{c|cccc}
            \hline \hline
            Method & BA & Physical ASR & IS & FID \\
            \hline
            Paste Directly & 95.04 & 43.80 & 1.00026 & $1.59 \times 10^{-2}$ \\
            Without Guidance & \textbf{97.03} & 57.92 & 1.00026 & $5.74 \times 10^{-3}$ \\
            DiffPhysBA & 95.16 & \textbf{69.40} & \textbf{1.00027} & $\bm{5.71 \times 10^{-3}}$ \\
            \hline\hline
        \end{tabular}
        \end{adjustbox}
        \label{subtable:ablation_guidance}
    \end{subtable}%
    \quad
    \begin{subtable}{0.45\textwidth}
        \centering
        \renewcommand{\arraystretch}{1.025} 
        \caption{The performance of different triggers.}
        \begin{adjustbox}{width=\textwidth}
        \begin{tabular}{c|ccccc}
             \hline \hline
            Trigger & BA & Digital ASR & SSIM & PSNR & LPIPS \\
            \hline
            Hats  & 96.01 & 91.20 & 0.0656 & 92.2 & 24.57 \\
            Shoes & \textbf{96.71} & 94.42 & 0.0417 & 92.13 & 23.33 \\
            Bags & 95.16 & \textbf{95.78} & \textbf{0.0321} & \textbf{93.42} & \textbf{25.72} \\ 
             \hline \hline
        \end{tabular}
        \end{adjustbox}
        \label{subtable:trigger_performance}
    \end{subtable}
    \vspace{-8mm}
\end{table}

\paragraph{Different trigger patterns.} The visualization experiments are illustrated in Figure \ref{trigger}, showcasing the stealthiness of various triggers, including hats and shoes. It indicates that hats and shoes may not seamlessly blend into the scene as the bag does. For quantitative results, please refer to Table \ref{subtable:trigger_performance}. They can be proven effective in the digital domain, with no notable distinction between bags. For reasons of stealthiness, we tend to use the bag as the trigger.
\section{Conclusion}
In this paper, we propose a diffusion-based physical backdoor attack (DiffPhysBA) against person ReID models. 
We present an innovative strategy that utilizes grounded diffusion models to create natural attributes within the digital domain, functioning as semantic-level triggers. 
Moreover, we propose a training-free ReID-driven similarity guidance for denoising steps, which enables the generated poisoned images to be closer to the original ones.
Our approach not only enhances the realism of the generated triggers but also facilitates a seamless transition from the digital to the physical domain, amplifying the potential impact of these triggers in practical attack scenarios.

\bibliographystyle{splncs04}
\bibliography{neurips}
\end{document}